\documentclass{article}

\usepackage[preprint]{neurips_2026}

\usepackage[utf8]{inputenc} 
\usepackage[T1]{fontenc}    
\usepackage{hyperref}       
\usepackage{url}            
\usepackage{booktabs}       
\usepackage{amsfonts}       
\usepackage{nicefrac}       
\usepackage{microtype}      
\usepackage{xcolor}         
\usepackage{amsmath}
\usepackage{graphicx}       
\usepackage{cleveref}
\usepackage{multirow}
\usepackage{subcaption}
\usepackage{bbm}
\usepackage{adjustbox}
\usepackage{microtype}
\usepackage{graphicx}
\usepackage{booktabs} 

\usepackage{hyperref}

\usepackage{amsmath}
\usepackage{amssymb}
\usepackage{mathtools}
\usepackage{amsthm}

\theoremstyle{plain}

\theoremstyle{definition}

\theoremstyle{remark}

\usepackage[textsize=tiny]{todonotes}

\usepackage{color, colortbl}
\definecolor{Gray}{gray}{0.93}
\definecolor{Orange}{rgb}{1,0.5,0}
\definecolor{DGray}{gray}{0.83}
\definecolor{LightCyan}{rgb}{0.88,1,1}
\usepackage[most, breakable]{tcolorbox}

\usepackage{bbm}
\usepackage{adjustbox}
\usepackage{lscape} 
\usepackage{makecell} 
\usepackage{bm}
\usepackage{float}
\usepackage{xspace}
\usepackage{nicematrix}
\usepackage{url}            
\usepackage{amsfonts}       
\usepackage{nicefrac}       
\usepackage{microtype}      
\usepackage{xcolor}         
\usepackage{pifont}

\usepackage{wrapfig}
\usepackage{multirow,makecell}
\usepackage{tabularx}
\usepackage{bbding}
\usepackage{threeparttable}
\usepackage{wasysym}
\usepackage{todonotes}
\usepackage{enumitem}
\usepackage{enumitem}
\setlist[itemize]{leftmargin=0.5cm}

\title{Stepped MoE: Segment-Level Routing with Configurable Inference Complexity}

\author{
Arnav Kundu\thanks{Corresponding author: arnav1993k@gmail.com}, Zhaoyang Xu, Bairu Hou, Chang Gao, Reed Li, Tao Lei\\
\texttt{Apple}
}

\begin{document}
\maketitle

\begin{abstract}
\vspace{-1mm}
Training large language models (LLMs) is resource-intensive, and adapting them for diverse deployment scenarios with varying computational constraints remains challenging. While elastic architectures enable flexible model deployment and sparsely activated models allow input-adaptive computation, existing approaches treat these dimensions independently. Moreover, models catered towards on-device edge inference need to conform to the memory and compute limitations of the serving devices. In this paper, we introduce a unified framework that combines elastic structures with sparsely gated architectures to create models that adapt simultaneously to both deployment constraints and task requirements. Our approach employs a model backbone that conditions on both the context and target efficiency specifications, enabling fine-grained control over the accuracy-efficiency trade-off at inference time. The model learns to activate task-relevant parameters within elastically-nested sub-networks, allowing a single model to span multiple capacity points while maintaining input-adaptive routing. Through experiments we demonstrate that we can create a model that allows the flexibility to use 1,2,3,4 billion parameters while being more accurate than their dense counter-parts (2-5\% on knowledge-intensive benchmarks) and at par with their static versions while delivering similar latency metrics as dense models. Overall, we save on device disk space by sharing the model parameters, allow flexibility of serving based on DRAM and compute available while delivering more accurate results.
\vspace{-2mm}
\end{abstract}

\vspace{-1mm}
\vspace{-2mm}
\section{Introduction}

The cost of running a large language models at inference time has become as much of a bottleneck as training it. One line of work that directly attacks this problem is sparse activation: instead of using every parameter for every input, the model learns to route each token to a small number of specialist sub-networks and leaves the rest untouched. Mixture of Experts (MoE)~\citep{shazeer2017outrageously,moe_expert_choice2022} is the best-known version of this idea. A lightweight, learned router examines each token in its surrounding context and picks the handful of expert modules most likely to be useful, while the remaining experts contribute nothing to that forward pass. This lets the model capacity grow substantially without a proportional jump in per-token FLOPs, which is why MoE models often match or beat their dense counterparts at a fraction of the serving cost. The arrangement is not free of trade-offs, though. Conventional MoE training enforces a fixed sparsity ratio at the granularity of individual tokens, so the specific subset of weights the accelerator needs changes at every single decoding step. This means the entire model should be loaded in memory for fast execution. For edge applications with typically smaller memory available, what should be a compute-bound workload becomes an I/O-bound one: the hardware spends most of its time shuffling expert weights between slow and fast memory rather than doing useful arithmetic.

Instruction Following Pruning (IFP)~\citep{hou2025instruction} addresses this by using a secondary model to select a sub-network once per user query. The selected parameters are then cached for the duration of the response, effectively reducing the problem to dense model inference at runtime. However, IFP introduces several practical challenges. First, training and maintaining a separate pruner model adds both training and deployment overhead. Second, IFP generates masks at the granularity of individual hidden dimensions, creating an extremely high-dimensional search space causing a training bottleneck because sparse-gem ops cannot be used. Additionally, the mask predictor size will astronomically increase. Third, and particularly relevant for on-device deployment, IFP offers no mechanism for users to trade inference speed against output quality at runtime. And finally, as per the IFP paper such a model cannot be pre-trained rather only upcycled.

This paper introduces a flexible Stepped-MoE architecture to extend the ideas of MoE and IFP into a segment based sparsely activated model that allows to adapt to inference platform capacity while allowing for a latency-accuracy tradeoff. Specifically, our contributions are
\begin{itemize}
\item Reducing the search space of the pruner in IFP by converting regular feed-forward blocks into MoE blocks, where the pruner select experts instead of individual rows and columns from the feed forward weights.
\item A unified routing mechanism that re-purposes some part of the model to act as the pruner, removing the need of a separate pruner model.
\item A stepped expert switching scheme where experts are reloaded every $S$ inference steps. This balances representational capacity, which suffers when sub-networks are fixed too rigidly as in IFP , against the latency cost of switching too frequently.
\item Configurable active parameter budgets drawn from a single trained model, allowing users to select the compute-accuracy tradeoff appropriate for their task and hardware constraints.
\end{itemize}
\begin{figure}
  \centering
  \includegraphics[width=1\linewidth]{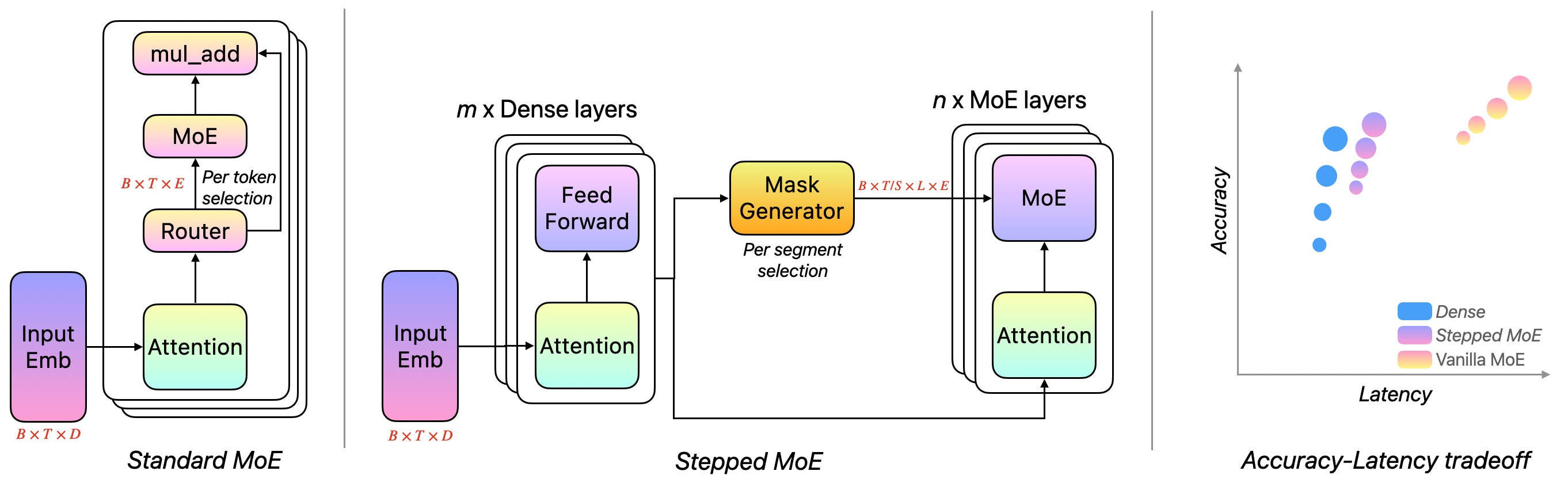}
  \caption{Left: Traditional MoE based architecture where experts are selected per token at each layer, Middle: Stepped MoE architecture that decides the experts for all layers to use at once using the mask generator, Right: Accuracy-Latency trade-off curve showing how stepped-MoE bridges the gap between Dense and MoE Models (size of the bubble represents active size of the models).}
  \label{fig:arch}
\end{figure}
As illustrated in \cref{fig:arch} the model is divided into 2 parts: the dense part and the MoE part. The mask generator uses the embeddings from the dense part to generate importance scores for every expert per layer in the MoE part. These scores are then sorted and top $k$ scoring indices are selected. The $k$ in this case is variable depending on the number of active parameters to be used. We ensure that the selection and ranking process is fully differentiable. The selected indices represent the indices of the experts in the corresponding MoE layers. We further introduce a sparsity-aware embedding that explicitly signals the model's active capacity, enabling more consistent behavior across different inference-time sparsity configurations. 

Since, the dense layers and the MoE layers along with the differentiable mask generation mechanism are part of the same model we can jointly optimize them by minimizing the next-token prediction loss. We employ effective training strategies that leverage both pre-training and supervised fine-tuning data. At test time, only the selected experts are used and refreshed after every $n$ tokens of generation.

We validate our model development process by running comprehensive set of pre-training and instruction finetuning experiments showing incremental gains of introducing each constraint. We also validate our results and demonstrate that we are able to create models that can serve under multiple sparsity constraints from the same total parameters while being more accurate that their dense counter-parts. We also demonstrate that the Stepped-MoE approach is much faster at inference time as compared to Vanilla MoE models. 
\section{Methodology}
\begin{figure}
  \centering
  \includegraphics[width=1\linewidth]{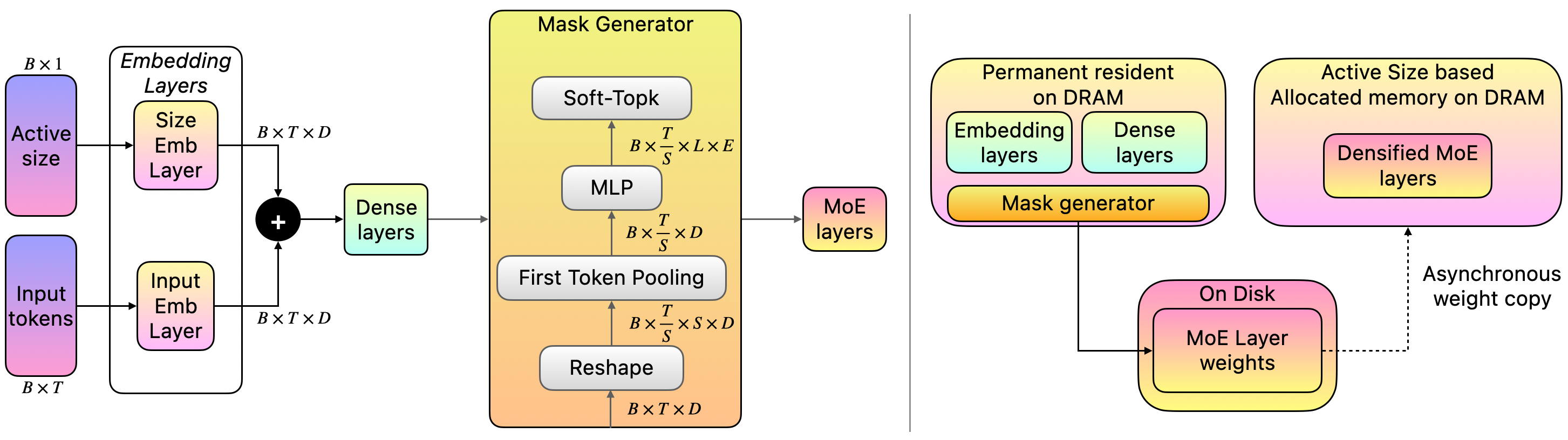}
  \caption{Left: Flexible Stepped-MoE model architecture with expanded mask generator, Right: Inference time layout of the model to allow flexible sparsity based on resource availability.}
  \label{fig:model}
\end{figure}
In this section, we elaborate on our model architecture, data mixture and training methodology. We focus primarily on the problems of building local LLMs that can operate under constrained resources while offering flexibility in terms of compute and accuracy. A large language model designed to operate under constraints of compute budgets needs to be flexible in allocating compute based on the users needs. Furthermore, when runtime memory (DRAM) available on users devices are limited the model needs to efficiently load and unload parameters from the disk without compromising on the speed of response generation.

\subsection{MoE background}
In Mixture of Experts (MoE) models~\citep{shazeer2017outrageously, moe_expert_choice2022}, every layer contains a router that ingests the previous layer's outputs and produces a routing mask over the experts in the current layer. The routing mask is governed by a parameter $k$ that caps the number of experts activated for a given input. Conventionally, routers compute a separate mask for each token, which limits the total FLOPs the model spends per token. Per-token routing, however, means that the set of active experts can change with every decoding step, forcing the system to evict some expert weights from DRAM and fetch replacements from disk each time a new token is generated. The constant swapping of expert weights hurts decoding throughput because typically disk-to-DRAM bandwidth is far lower than the arithmetic throughput of current-generation accelerators. To avoid this I/O penalty in practice, serving systems typically keep the entire model resident in DRAM, even though only a small fraction of those parameters are actually used at any single inference step. This limits the applicability of such models on DRAM constrained devices.

\subsection{IFP background}
Instruction Following Pruning (IFP)~\citep{hou2025instruction} introduced a sparsity predictor, a tiny neural network that produces a pruning mask conditioned on a given query. During response generation, only the subset of parameters selected by the pruning mask is loaded into DRAM. Those weights remain resident for the entire duration of that query. Since the active parameter set is fixed per query rather than changing at every token, IFP avoids the repeated expert-swapping bottleneck that plagues conventional MoE serving. However, committing to a single set of experts for an entire query also has downsides: the model cannot adapt its active parameters as the context evolves during generation, which can hurt output quality. Additionally, IFP treats each hidden dimension of the feed-forward layer as an expert, therefore has a massive optimization space with such high granularity of experts.

\subsection{Model architecture}
Our base model architecture: Stepped MoE is inspired by MoE~\citep{shazeer2017outrageously} and IFP~\citep{hou2025instruction} and tries to take advantage of both the methods catered to on-device inference. The architecture is shown in \cref{fig:model}. Contrary to IFP our model doesn't have an explicit pruner model. Instead we use the initial layers of a dense transformer decoder model as the backbone for the expert selector. Since experts for all successive layers are selected at once, parameters for the later layers can be loaded independently without waiting for each layer to be executed. In other words, the early decision of which experts to load helps to parallelize the compute and IO by loading future experts while initial layers are being executed. We avoid per token expert load and flush needed to serve MoE models with limited DRAM by ensuring that the experts selected stay valid for a fixed segment length. 

The expert selector takes outputs from the dense transformer backbone $(z\in nB\times T \times D )$ and maps it to routing logits for every successive layers in the model $(g\in nB\times \frac{T}{S} \times L \times E )$. These routing logits are then passed through a SoftTopK~\citep{lei2023conditional,ainslie2023colt5} function to generate expert masks $(M\in\{0,1\})$ to be used by the MoE layers. This mask is refreshed after every $S$ tokens.
The MoE layers act similar to general token based MoE models, the only difference being that the experts are pre-selected and the outputs from the experts are not combined using routing weights. The size embedding $W_k$ is a learned vector indexed by active budget $k_i$, added to the token embeddings before the dense layers $h_0=\text{TokenEmb}(x)+W_{k_i}$. This gives the model an explicit signal about the target sparsity regime, which we find critical for consistent routing behavior across budget levels.

\subsection{Training recipe}
\subsubsection{Pre-training}
Our pre-training recipe caters to two important aspects of our model architecture: segment-wise expert selection and supporting multiple sparsities. At pre-training stage the dense part of the model ingests input tokens of length $T$ to generate an embedding of the same length. The pruner divides this embedding into chunks of segment length $S$. We use the first embedding in each of these segments to decide the expert selection mask for the segment for all the following MoE layers. Contrary to~\citep{hou2025instruction} this segment wise mask selection allows us to pre-train a model from scratch.

For typical MoE based models the number of experts selected ($k$) is an integer, therefore making it a single sparsity model. We wanted our model to be able to work under a list of sparsities. We define a vector ($k$) to represent the required sparsity per batch element. This vector can be randomly populated with required sparsity levels or to make the training more deterministic we can use a static pattern. For the sake of simplicity, we repeat the list of supported sparsities to fill a vector of the same size as the input batch size. We also provide the same vector as input to the active size embedding table. We introduce batch replication by repeating each batch element $n$ times making our effective batch size $n$ times larger. In this manner each batch sample is trained for all supported active sizes.

\subsubsection{Loss formulation}
For a given batch the loss computed over variable active parameters. The loss computed can be represented as follows:
\begin{equation}
  L = \frac{1}{n\times B}\Sigma_i^n\Sigma_{j=i\times B} ^{(i+1)\times B} \text{CE}(y_j, f(W_i,x_j)).
\end{equation}

Here, $W_i$ corresponds to the set of weights selected for a $i^{th}$ sparsity level. The exact weights in $W_i$ might be different across samples for the given sparsity level. $x_j, y_j$ corresponds to the $j^{th}$ input and output pair for a given sparsity. $B$ and $n$ correspond to the per sparsity batch size and number of sparsities available respectively. Lets assume, B=1 and n=2, we can write the loss function as
\begin{equation}
  L = \frac{1}{2} (\text{CE}(y_1, f(W_1,x_1)) +\text{CE}(y_2, f(W_2,x_2))),
\end{equation}
where $W_1$ corresponds to the most sparse configuration and $W_2$ is the configuration with most active parameters. Therefore, $W_1\subset W_2$ and we can write $W_2 = [W_1,W_e]$, where $W_e = W_2-(W_1\cap W_2)$. Subsequently, the gradient of $W_1$ is a function of both the inputs therefore, is updated more often, whereas the gradient of $W_e$ is sparser therefore needs stronger update signals. For balancing this difference in gradient updates for less sparse configurations in our multi-sparse model we introduce a sparsity based loss scaling that penalizes denser models more than sparser models as follows:

\begin{equation}
  L = \frac{1}{n\times B}\Sigma_i^n\alpha_i\times \Sigma_{j=i\times B} ^{(i+1)\times B} \text{CE}(y_j, f(W_i,x_j)),
\end{equation}
where $\alpha_i$ corresponds to the scaling factor used for sparsity level $i$.
We demonstrate the contribution of this loss function through our experiments in the \cref{tab: ablt_exp} in \cref{sec:Experiments}. We acknowledge that the mathematically correct formulation to introduce the desired effect would have been to have different learning rates for experts that are activated less frequently. However such an implementation is much more complex therefore we use the loss scaling as a simpler proxy.
\subsubsection{Supervised Fine-tuning}
SFT follows the pretraining setup, with the following differences:
\begin{itemize}
\item Multiple short examples are concatenated up to the max sequence length.
\item The loss is computed only on {\it target tokens} (generally assistant turn tokens, not user prompt tokens)
\item We use the last prompt token to select experts for all prompt tokens and the first response segment. For the remaining segments, we use the first token in each segment to maintain the causality.
\end{itemize}
\subsection{Inference}
During inference, as illustrated in \cref{fig:model} the dense part of the model is permanently loaded on the DRAM for all model serving use cases. Based on the task and device capacity an active size is selected by the user. Based on the selected active size, memory is allocated on DRAM corresponding to dense parameters' size of the MoE layers. For every input prompt, the first set of experts for response generation are selected based on the last token of the prompt. These experts are valid for the segment length $S$ generation steps and refreshed every $S$ steps from there on. Since, the experts for all the layers are selected at once, experts for later layers can be asynchronously loaded from disk while earlier experts are getting executed.
\section{Experiments}
\label{sec:Experiments}
\subsection{Pre-training}
Our model has 56 transformer layers out of which the first 12 are dense layers that also help to generate the expert routing mask for the later 44 layers. We use 1536 as our base model dimension. The attention layers use grouped query attention with 12 query and KV heads. The dense part of the model use 2x scaling for the feed forward layers while the MoE layers have 212 experts where expert hidden dimension is 256. These architectural hyper-parameters are a result of small scale experiments on static S-MoE models and might change on scaling to different model configurations.

For pre-training, we use general continuous stream of input text tokens collected from web-scraped data. The models are trained over 1 trillion tokens with a context length of 8192 and batch size of 2048 which is replicated 4 times to serve each level of sparsity. We use decoupled weight decay for regularization along with a simplified $\mu Param$ optimizer. Our best training run uses a weighted cross-entropy loss where the weight is calculated as a function of number of routed experts per sample.
We use the AXLearn~\citep{apple2023axlearn} framework and JAX~\citep{bradbury2018jax} for model training. 

In our experiments, we establish Dense baselines ranging from 1-4B parameters trained on the same dataset. Additionally, we also report metrics of Static Stepped-MoE models to demonstrate the accuracy of independently trained Stepped-MoE models. Finally, we demonstrate the results of the proposed Flexible Stepped-MoE models across 4 levels of active parameters. We also establish a 3B active 12B total parameter Vanilla MoE model as the upper bound for accuracy. At pre-training stage we evaluate the models on the following datasets:
\begin{itemize}
\item Core Text. We include a set of tasks to evaluate the model's core capabilities of natural language understanding, scientific knowledge, and reasoning.
 We report the zero-shot performance on average of \texttt{ARC-challenge}~\citep{clark2018think}, \texttt{ARC-easy}~\citep{clark2018think}, \texttt{HellaSwag}~\citep{zellers2019hellaswag}, \texttt{WinoGrande}~\citep{sakaguchi2021winogrande}, \texttt{PiQA}~\citep{Bisk2020}, \texttt{LAMBADA-OpenAI}~\citep{paperno2016lambada}, and \texttt{SciQ}~\citep{welbl2017crowdsourcing}.We also report one-shot performance on average of \texttt{Trivia-QA}~\citep{joshi2017triviaqa} and \texttt{Web-QS}~\citep{webqs}.
 
\item MMLU~\citep{hendrycksmeasuring}. We evaluate the 5-shot performance and report the multiple-choice accuracy.

\item Math. We use \texttt{GSM8K}~\citep{cobbe2021training} to evaluate the math capabilities of LLMs. We report the 8-shot exact-match accuracy.
\item Coding tasks. We evaluate the pass@1 performance on \texttt{HumanEval-python}~\citep{chen2021codex}.
\end{itemize}
\newcommand{\resoneb}[1]{\textcolor{red}{#1}}
\newcommand{\restwob}[1]{\textcolor{blue}{#1}}
\newcommand{\resthreeb}[1]{\textcolor{teal}{#1}}
\newcommand{\resfourb}[1]{\textcolor{brown}{#1}}
\begin{table*}[ht]
\centering
\caption{Pre-training results; results for the same active parameters are same color coded and best results are in bold.} 
\vspace{0.1in}
\begin{adjustbox}{max width=.88\textwidth}
\begin{tabular}{c|c|c|c|c|c|c}
\toprule
\multirow{2}{*}{\textbf{Model Type}} & \multirow{2}{*}{\textbf{\,\,\,\,Active Parameters}} & \multirow{2}{*}{\makecell{Core-En-0s}} & \multirow{2}{*}{\makecell{Core-En-1s}} & \multirow{2}{*}{\makecell{MMLU-5s}} & \multirow{2}{*}{\makecell{GSM-8k}} & \multirow{2}{*}{\makecell{humaneval-py}}\\
& & & & & \\
\midrule
\multirow{4}{*}{\makecell{\textbf{Dense}}}
& \resoneb{1B} & \resoneb{\textbf{0.662}} & \resoneb{0.19} & \resoneb{0.470} & \resoneb{0.174} & \resoneb{0.132} \\
& \restwob{2B} & \restwob{\textbf{0.690}} & \restwob{0.251} & \restwob{0.552} & \restwob{0.260} & \restwob{0.156} \\
& \resthreeb{3B} & \resthreeb{0.704} & \resthreeb{0.282} & \resthreeb{0.578} & \resthreeb{0.345} & \resthreeb{0.184} \\
& \resfourb{4B} & \resfourb{\textbf{0.716}} & \resfourb{\textbf{0.296}} & \resfourb{0.610} & \resfourb{0.411} & \resfourb{0.205} \\
\midrule
\multirow{2}{*}{\makecell{\textbf{Static S-MoE}}}
& \resoneb{1B} & \resoneb{0.652} & \resoneb{\textbf{0.212}} & \resoneb{0.544} & \resoneb{0.186} & \resoneb{0.156} \\
& \resthreeb{3B} & \resthreeb{0.704} & \resthreeb{0.264} & \resthreeb{0.621} & \resthreeb{0.460} & \resthreeb{0.264} \\
\midrule
\multirow{4}{*}{\makecell{\textbf{Flexible S-MoE}}}
& \resoneb{1B} & \resoneb{0.640} & \resoneb{0.196} & \resoneb{\textbf{0.5706}} & \resoneb{\textbf{0.283}} & \resoneb{\textbf{0.170}} \\
& \restwob{2B} & \restwob{0.683} & \restwob{\textbf{0.259}} & \restwob{\textbf{0.6144}} & \restwob{\textbf{0.459}} & \restwob{\textbf{0.243}} \\
& \resthreeb{3B} & \resthreeb{0.701} & \resthreeb{0.279} & \resthreeb{0.6227} & \resthreeb{\textbf{0.480}} & \resthreeb{0.243} \\
& \resfourb{4B} & \resfourb{0.712} & \resfourb{0.29} & \resfourb{\textbf{0.6317}} & \resfourb{\textbf{0.483}} & \resfourb{\textbf{0.248}} \\
\midrule
\multirow{1}{*}{\makecell{\textbf{Vanilla MoE}}}
& \resthreeb{3B} & \resthreeb{\textbf{0.725}} & \resthreeb{\textbf{0.305}} & \resthreeb{\textbf{0.662}} & \resthreeb{\textbf{0.528}} & \resthreeb{\textbf{0.278}} \\
\bottomrule
\end{tabular}
\end{adjustbox}
\label{tab: main_exp}
\end{table*}
It can be observed that, across most sizes flexible S-MoE outperforms its dense counter-parts in almost all benchmarks and is worse than Vanilla MoE which uses per token expert swaps. As discussed earlier, we intend to hit a trade-off curve between latency and accuracy using Stepped MoE and Vanilla MoE is the upper-bound for accuracy in this curve. For some sizes like 1B in some benchmarks like Core-en-0s the accuracy of stepped MoE seems to be worse than its dense counterpart but overall in knowledge based tasks like MMLU the gains seem to be much significant. We hypothesize the 12 dense overhead layers represent a disproportionately large fixed cost at 1B scale (21\% of all layers), leaving insufficient capacity for MoE-specific learning.

\subsection{Pre-training Ablations}
We have introduced a number of ideas to make sure our flexible sparsity model is atleast as accurate as its dense and static counterparts. We demonstrate the impact of each of those steps. When we go from Static S-MoE to Flexible S-MoE, we observe that the performance of the model drops especially for the higher activated sizes. This can be attributed to the optimization problem being more difficult for the later since it has to train 4 different models with the same shared parameters. If we account for the training FLOPs equivalent of all the extractable models from Flexible Stepped-MoE, it comes out to be approximately same as training 4 dense models. Therefore, we believe training Flexible Stepped-MoE for the same training FLOPs as a dense model leaves it under-trained and we introduce batch replication where every input sequence in a batch is replicated 4 times to resolve this gap. Matformer~\citep{kudugunta2023matformer} observes a similar trend in scaling compute as more sizes are introduced. We primarily use batch replication instead of training for longer to ensure that the data seen by all the models in these ablations are same. Additionally, we demonstrate that introducing mask weighted CE Loss improves the larger models while reducing the performance of small active size model as compared to the static baselines. Finally, the active size embedding helps to bridge this gap by introducing a signal for the model to know the size constraint and accordingly adjust its parameters.
\begin{table*}[ht]
\centering
\caption{Ablations showing the impact of incremental changes to make Flexible S-MoE work}
\vspace{0.1in}
\begin{adjustbox}{max width=.88\textwidth}
\begin{tabular}{c|c|c|c|c|c}
\toprule
\multirow{2}{*}{\textbf{Model Type}} & \multirow{2}{*}{\textbf{\,\,\,\,Active Parameters}} & \multirow{2}{*}{\makecell{Core-En-0s}} & \multirow{2}{*}{\makecell{Core-En-1s}} & \multirow{2}{*}{\makecell{MMLU-5s}} & \multirow{2}{*}{\makecell{GSM-8k}} \\
& & & & & \\
\midrule
\multirow{2}{*}{\makecell{\textbf{Dense}}}
& 1B & 0.662 & 0.19 & 0.470 & 0.174 \\
& 3B & 0.704 & 0.282 & 0.578 & 0.345 \\
\midrule
\multirow{2}{*}{\makecell{\textbf{Static S-MoE}}}
& 1B & 0.652 & 0.212 & 0.544	& 0.186	\\
& 3B & 0.704 & 0.264 & 0.621 & 0.460 \\
\midrule
\multirow{2}{*}{\makecell{\textbf{Flexible S-MoE}}}
& 1B & 0.650 & 0.204 &0.500 &	0.200 \\
& 3B & 0.686 & 0.254 & 0.584 &	0.405 \\
\midrule
\multirow{2}{*}{\makecell{\textbf{+ Batch Replication}}}
& 1B &0.657 & 0.202 &	0.576&	0.304\\
& 3B &0.698 & 0.264 &	0.598&	0.414\\
\midrule
\multirow{2}{*}{\makecell{\textbf{+ Weighted CE}}}
& 1B &0.635 & 0.184 &	0.556&	0.277\\
& 3B &0.699 & 0.254 &	0.617&	0.459\\
\midrule
\multirow{2}{*}{\makecell{\textbf{+ Active Size Embedding}}}
& 1B &0.640 & 0.196 & 0.5706 & 0.283\\
& 3B &0.701 & 0.279 & 0.6227 & 0.480 \\
\bottomrule
\end{tabular}
\end{adjustbox}
\label{tab: ablt_exp}
\end{table*}

We further ablate on the choice of segment length, i.e., the number of tokens over which expert assignments remain static. A shorter segment length allows the model to re-route tokens to different experts more frequently, offering finer-grained adaptability to the input. However, this comes with a cost of increased routing overhead at inference time. Conversely, a longer segment length reduces routing frequency, reducing serving latency, but may limit the model's ability to dynamically adapt expert selection to evolving context within a sequence. As shown in Table~\ref{tab: ablt_seg}, a segment length of 16 outperforms a segment length of 32 across all benchmarks, suggesting that the gains from finer-grained routing outweigh the associated costs in this regime. This observation is coherent with the fact that vanilla MoE outperforms S-MoE in all benchmarks because of per-token routing. The optimal segment length is likely to depend on factors such as sequence length, model scale, and deployment constraints, and we treat it as a configurable hyperparameter rather than a fixed design choice.
\begin{table*}[ht]
\centering
\caption{Ablations showing the impact of segment length}
\vspace{0.1in}
\begin{adjustbox}{max width=.88\textwidth}
\begin{tabular}{c|c|c|c|c|c}
\toprule
\multirow{2}{*}{\textbf{Model Type}} & \multirow{2}{*}{\textbf{\,\,\,\,Active Parameters}} & \multirow{2}{*}{\makecell{Core-En-0s}} & \multirow{2}{*}{\makecell{Core-En-1s}} & \multirow{2}{*}{\makecell{MMLU-5s}} & \multirow{2}{*}{\makecell{GSM-8k}} \\
& & & & & \\
\midrule
\multirow{2}{*}{\makecell{\textbf{Flexible S-MoE} \\ \textbf{Segment length=32}}}
& 1B &0.640 & 0.196 & 0.5706 & 0.283 \\
& 3B &0.701 & 0.279 & 0.6227 & 0.480 \\
\midrule
\multirow{2}{*}{\makecell{\textbf{Flexible S-MoE} \\ \textbf{Segment length=16}}}
& 1B & 0.666 & 0.223 & 0.601 & 0.305 \\
& 3B & 0.722 & 0.298 &	0.653 & 0.513 \\
\bottomrule
\end{tabular}
\end{adjustbox}
\label{tab: ablt_seg}
\end{table*}

\subsection{Supervised fine-tuning}
After pre-training, we use supervised fine-tuning (SFT) datasets to instruction tune our model. The SFT training for the baselines is performed with a batch size of 512 for 30k training steps. For Flexible S-MoE we use batch size of 128 with 4x batch replication. We use the last prompt token to select experts for the prompt tokens as well as the first segment. From there on experts are reselected for every segment.
The SFT data mixture composition is as follows: 35\% instruction following, 20\% multilingual, 20\% math, 10\% coding, 10\% tool use, 5\% safety and others. We originally tuned all the hyper-parameters for the dense 3B model and adopted it for all experiments. SFT uses a training sequence length of 16k tokens.

For evaluations we use the following datasets:
\begin{itemize}[leftmargin=10pt,itemsep=2pt, topsep=2pt]
  \item Instruction-following. We include \texttt{IFEval}~ ~\cite{zhou2023instruction}, \texttt{AlpacaEval 2.0}~ ~\cite{dubois2024length}, and \texttt{Arena-Hard-Auto}~ ~\cite{li2024crowdsourced} for evaluation. We report the prompt-level and instruction-level accuracy on IFEval, the length-controlled win rate on AlpacaEval 2.0, and win rate on Arena-Hard-Auto.
  \item Coding tasks. We evaluate the pass@1 performance on \texttt{HumanEval-python}~ ~\cite{chen2021codex}, \texttt{MBPP}~ ~\cite{austin2021program}, and \texttt{MultiPL-E}~ ~\cite{cassano2022multipl}. For MultiPL-E benchmark, we report the average metric over Swift and Java metrics.
  \item Math. We use \texttt{GSM8K}~ ~\cite{cobbe2021training} and \texttt{COT-Minerva}~ ~\cite{hendrycksmath2021} to evaluate the math capabilities of LLMs. we report the accuracy with few-shot examples on both datasets (8-shot for GSM8K and 4-shot for MATH).
  \item Tool use. We evaluate the tool use performance on \texttt{MMAU}~ ~\cite{yin2024mmau} and report the performance on Tool Execution.
\end{itemize}

\begin{table*}[ht]
\centering
\caption{Performance comparison between Dense models, Static Stepped-MoE (S-MoE) and Flexible Stepped MoE (S-MoE). All S-MoE models have 12B total parameters.
}
\vspace{0.1in}
\begin{adjustbox}{max width=.95\textwidth}
\begin{tabular}{c p{3cm}|cc|c|cccc}
\toprule
\multirow{2}{*}{\textbf{Category}} & \multirow{2}{*}{\textbf{\,\,\,\,Dataset}} & \multirow{2}{*}{\makecell{\textsc{Dense} 3B}} & \multirow{2}{*}{\makecell{\textsc{Dense} 6B}} & {\makecell{Static S-MoE}} & \multicolumn{4}{c}{{Flexible S-MoE}} \\
& & & & 12B$\rightarrow$3B & 12B$\rightarrow$1B & 12B$\rightarrow$2B & 12B$\rightarrow$3B & 12B$\rightarrow$4B\\
\midrule
\multirow{4}{*}{\makecell{\textbf{Instruction}\\ \textbf{Following}}}
& {IFEval-Instruction}  & 83.693	& 86.451 & 83.813 & 76.739 &	82.494	& 85.252	& 85.372
 \\
& {IFEval-Prompt}  & 76.895 & 79.852 & 77.449 & 66.543 & 74.677 &	79.482 & 78.743
\\
& {AlpacaEval-Prompt}  & 16.479 &	27.333 & 22.951 & 13.597 &	20.409	& 22.111	& 25.021
\\
& {Arena-Hard}  & 10.85 &	19.49 & 19.07 & 10.95 &	17.72	& 19.75	& 18.49
\\
\midrule
\multirow{1}{*}{\textbf{Tool Use}}
& {Tool Execution} & 75.736	& 80.102 & 77.97 & 67.614 & 76.244 & 77.056 & 78.071 \\
\midrule
\multirow{2}{*}{\textbf{Math}}
&	COT-Minerva &  36.900	& 	46.229	& 	43.514	& 	33.657	& 	42.943	& 	43.414	& 	45.071 \\
&	GSM-8k	&	64.600	& 	73.800	& 	70.000	& 	52.200	& 	68.100	& 	72.200	& 	74.100 \\
\midrule
\multirow{3}{*}{\textbf{Coding}}
&  MBPP  &	38.300	& 	45.900	& 	45.600	& 	35.300	& 	46.200	& 	46.200	& 	46.500	\\
&  HumanEval  &	35.600	& 	48.400	& 	44.250	& 	34.600	& 	44.550	& 	45.700	& 	46.100	\\
&  MultiPL-E  &	24.150	& 	29.900	& 	34.450	& 	21.500	& 	32.550	& 	32.900	& 	32.950	\\
\midrule
\bottomrule
\end{tabular}
\end{adjustbox}
\label{tab: sft_exp}
\end{table*}

\subsection{Inference profiling}

Stepped-MoE architecture intends to solve the drawbacks of MoE architecture for limited DRAM on-device use cases. We compare the performance of Stepped-MoE with 3B and 6B dense models as the target latency benchmark. For profiling, we run the model through a neural processing unit (NPU) simulator to ensure deterministic compute latencies for defined FLOPs. We use true matmul latencies and NAND to DRAM memory transfer times. For the dense models there is no NAND to DRAM transfer once the models are loaded while for Stepped-MoE and Vanilla MoE models experts are loaded and flushed from DRAM based on the experts selected.
\begin{figure}
    \centering
    \includegraphics[width=1\linewidth]{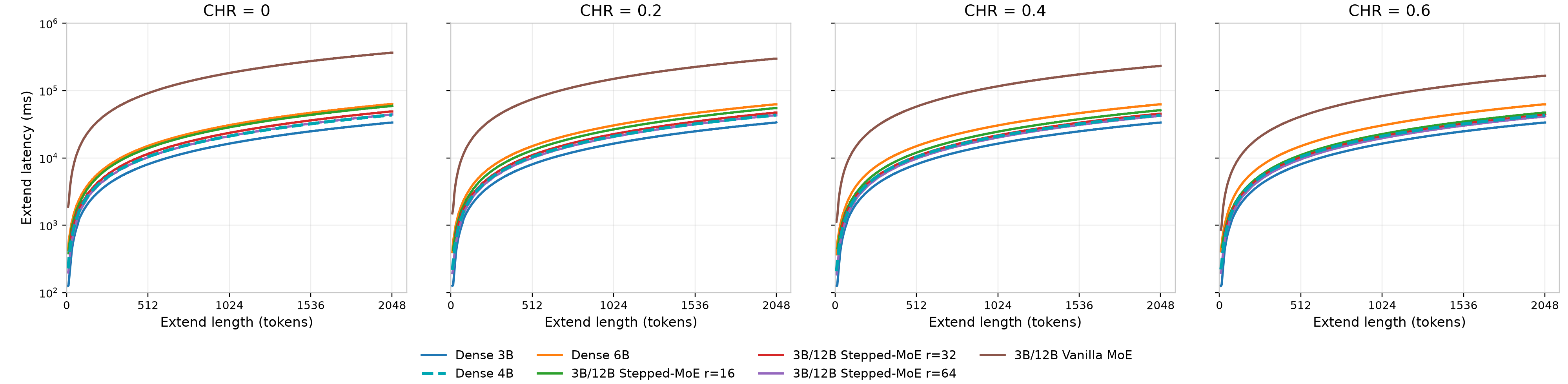}
    \caption{Decoding length vs Latency of various models and CHR: Cache Hit Rate; ratio of experts already present in DRAM from previous decoding step. Left $\Rightarrow$ Right we vary CHR from 0 to 0.6 to show the latency decreasing for Vanilla and Stepped-MoE models}
    \label{fig:latency}
\end{figure}
As shown in Figure \ref{fig:latency}, we evaluate the latency across different Cache Hit Rates (CHR): the ratio of experts already present in DRAM from the previous decoding step. As CHR increases from 0 to 0.6 (left to right), the latency of the Vanilla MoE naturally decreases but remains exponentially higher than the dense baselines due to continuous expert swapping. In contrast, Stepped-MoE maintains a nearly flat and stable latency profile across all sequence lengths. Regardless of the CHR, Stepped-MoE closely tracks the latency of the 3B Dense model and remains strictly faster than the 6B Dense model for different segment lengths ($r$), proving that segment-level routing successfully mitigates the I/O bottlenecks of traditional per-token routing. This can be primarily attributed to determination of all experts at once and switching experts less frequently. It also closely matches the latency of the 4B dense model while being more accurate than it.
\section{Related Works}
Mixture-of-Experts (MoE) architectures have emerged to be a powerful method for scaling language models while keeping inference compute tractable. ~\citep{shazeer2017outrageously} introduced sparsely-gated MoE layers, demonstrating that routing tokens to a subset of specialized experts enables significant capacity gains without proportional compute increases. DeepSeekMoE~\citep{dai2024deepseekmoe} further refined the MoE design by introducing shared experts that are always active alongside routed experts, a design we adopt in our architecture to ensure a stable baseline capacity across all sparsity configurations. A fundamental limitation shared by the above approaches is that routing decisions are made per-token at every layer, requiring experts to be swapped in and out of DRAM at each decoding step. This makes serving speed IO-bound, particularly on memory-constrained devices. Additionally, these models are trained to operate in a fixed sparsity mode determined during training. Therefore, the computational footprint of these models is not flexible and cannot be adapted to various platform constraints. IFP~\citep{hou2025instruction} solved this by performing instruction based pruning of the model using a separate pruner model. While solving the inference bottleneck the authors found the requirement of a separate pruner model and the inability to pre-train such a model from scratch as limiting factors.

Flexible complexity models focuses on building models that can flexibly adjust their computational footprint at inference time. Slimmable Neural Networks~\citep{yu2018slimmable} introduced the concept of elastic width, training a single convolutional network that can execute at multiple channel widths by sharing parameters across capacity levels. Once-for-All~\citep{cai2019once} extended this idea to jointly train a network that supports diverse architectural configurations — varying depth, width, and kernel size, enabling deployment across heterogeneous hardware without retraining. DynaBERT~\citep{hou2020dynabertdynamicbertadaptive} brought elastic inference to transformer models, allowing adaptive width and depth selection for BERT through structured knowledge distillation, demonstrating that a single fine-tuned model can serve multiple accuracy-efficiency operating points. Matformer~\citep{kudugunta2023matformer} introduced nested sub-structures inside transformer feed-forward blocks to allow for flexible sizes and demonstrated steps to train such models. These works establish the core principle we build upon: that a single set of shared parameters can support multiple sub-networks of varying capacity. Our approach extends this to MoE models with input-adaptive routing, and further introduces explicit sparsity-aware embeddings to signal the active capacity configuration to the model at inference time, which we find is critical for consistent behavior across sparsity levels.
\section{Conclusion and Limitations}
\label{sec:conclusion}
The two most popular ways to make LLMs cheaper at inference time, ie. sparse MoE routing and query-level pruning, each solve half the problem of constrained inference. Per-token MoE routing keeps the model adaptive but turns serving into an I/O bottleneck on memory-constrained devices. Query-level pruning like IFP fixes the I/O problem but locks the model into a single sub-network for the entire response. We developed Flexible Stepped-MoE to build compute adaptable models with high throughput on low resource devices.

Our results indicate 3B and 4B active configurations consistently match or beat their dense counterparts, and in several instruction-following and math tasks the Flexible model outperforms the equivalent Static S-MoE as well. This simplifies the technical debt of training and deploying multiple model using a single 12B-parameter checkpoint. On the inference side, the NPU profiling results validate that stepped-MoE is a better alternative than Vanilla-MoE models for on device use cases primarily because it is almost as fast as equally compute intensive dense models while being far better in accuracy. Our ablations only scratch the surface of showing that shorter segments improve accuracy but increase I/O pressure, and the right operating point will depend heavily on the target hardware's memory bandwidth characteristics.

We acknowledge that we were not able to demonstrate the consequences of training these models longer where we anticipate a larger gap between vanilla-MoE and stepped-MoE models. Furthermore, the attention parameters in our model serve all the active sizes which might saturate the accuracy of the larger active size models when trained longer. A possible next direction to solving this bottleneck might be through early exiting for smaller active size models. In future, we would also try to adopt the model to support variable segment length based on context. We would also explore a larger total and active parameter models to observe the scaling trend of this approach and examine the viability of this method for server size models. This can alleviate the ever increasing demand of DRAM for serving infrastructure. Finally, we acknowledge the pre-training compute used for training the flexible size model is almost equal to training equivalent static stepped-MoE models and we will work towards optimizing the recipe to minimize the training compute required.


\bibliographystyle{unsrtnat}
\bibliography{references}





\newpage
\appendix

\section{Technical Appendices and Supplementary Material}
\subsection{Impact Statement}\label{impact}
In this paper, our primary goal is to develop an algorithm to train a single model to serve at multiple active parameter sizes on a resource constrained device. Our method is designed to improve both inference efficiency and the performance of LLMs. The training data has been carefully filtered to ensure quality and safety; for instance, all instances of personal data in the SFT data were removed to uphold privacy standards. All the data collection process strictly adheres to ethical guidelines for data use, ensuring that no private or sensitive information is included in the training or evaluation process.

Also, the algorithm proposed in this work does not introduce additional risks of bias or harm with the underlying large language model. Furthermore, our method enhances computational efficiency, potentially reducing the environmental impact of large-scale model inference. While acknowledging that any machine learning model has the potential for misuse, we focus on safe and task-specific applications, such as math, coding, and tool use. We encourage further research into mitigating biases and unintended consequences in language models and remain committed to the responsible and ethical advancement of AI technologies.
\subsection{Scalablity of Stepped-MoE}
\begin{table}[ht]
\centering
\caption{Pre-training results for larger total parameters S-MoE models}
\vspace{0.1in}
\begin{adjustbox}{max width=.88\textwidth}
\begin{tabular}{c|c|c|c|c|c|c}
\toprule
\multirow{2}{*}{\textbf{Model Type}} & \multirow{2}{*}{\makecell{\textbf{\,\,\,\,Active Parameters /}\\ \textbf{Total parameters}}} & \multirow{2}{*}{\makecell{Core-En-0s}} & \multirow{2}{*}{\makecell{Core-En-1s}} & \multirow{2}{*}{\makecell{MMLU-5s}} & \multirow{2}{*}{\makecell{GSM-8k}} & \multirow{2}{*}{\makecell{humaneval-py}}\\
& & & & & \\
\midrule
\multirow{2}{*}{\makecell{\textbf{Static S-MoE}}}
& 3B/12B & 0.704 & 0.264 & 0.621 & 0.460 & 0.264 \\
& 3B/20B & 0.703 & 0.27 & 0.634	& 0.508 & 0.260 \\
\midrule
\midrule
\multirow{8}{*}{\makecell{\textbf{Flexible S-MoE} }}
& 1B/12B & 0.640 & 0.196 & 0.5706 & 0.283 & 0.170 \\
& 2B/12B & 0.683 & 0.259 & 0.6144 & 0.459 & 0.243 \\
& 3B/12B & 0.701 & 0.279 & 0.6227 & 0.480 & 0.243 \\
& 4B/12B & 0.712 & 0.29 & 0.6317 & 0.483 & 0.248 \\
\cmidrule{2-7}
& 1B/20B & 0.648	& 	0.2185	& 	0.591	& 	0.317	& 	0.178 \\
& 2B/20B & 0.687	& 	0.2725	& 	0.634	& 	0.478	& 	0.247 \\
& 3B/20B & 0.701	& 	0.295	& 	0.646	& 	0.507	& 	0.256 \\
& 4B/20B & 0.705	& 	0.307	& 	0.65	& 	0.517	& 	0.255 \\
\bottomrule
\end{tabular}
\end{adjustbox}
\label{tab: appendix_scalability_exp}
\end{table}
We investigate the effect of increasing the total parameters while keeping the active parameters same and find that larger total parameter count helps to improve accuracy in most benchmarks.
\subsection{Batch replication vs more training data}
\begin{table*}[ht]
\centering
\caption{Pre-training results for batch replication vs more training data with no batch replication}
\vspace{0.1in}
\begin{adjustbox}{max width=.88\textwidth}
\begin{tabular}{c|c|c|c|c|c}
\toprule
\multirow{2}{*}{\textbf{Model Type}} & \multirow{2}{*}{\makecell{\textbf{\,\,\,\,Active Parameters}}} & \multirow{2}{*}{\makecell{Core-En-0s}} & \multirow{2}{*}{\makecell{Core-En-1s}} & \multirow{2}{*}{\makecell{MMLU-5s}} & \multirow{2}{*}{\makecell{GSM-8k}}\\
& & & & \\
\midrule
\multirow{8}{*}{\makecell{\textbf{Flexible S-MoE} }}
& 1B & 0.640 & 0.196 & 0.570 & 0.283 \\
& 2B & 0.683 & 0.259 & 0.614 & 0.459 \\
& 3B & 0.701 & 0.279 & 0.623 & 0.480 \\
& 4B & 0.712 & 0.290 & 0.632 & 0.483 \\
\cmidrule{2-6}
& 1B & 0.633	& 	0.190	& 	0.573	& 	0.296	\\
& 2B & 0.690	& 	0.251	& 	0.637	& 	0.486	\\
& 3B & 0.710	& 	0.276	& 	0.648	& 	0.528	\\
& 4B & 0.719	& 	0.291	& 	0.652	& 	0.546	\\
\bottomrule
\end{tabular}
\end{adjustbox}
\label{tab: appendix_batch_replication}
\end{table*}
We investigate the effect of batch replication vs using more unique training tokens and find that on smaller active sizes do not improve accuracy significantly while the extra unique tokens improve MMLU performance on larger active sizes. This is most likely due to presence of novel information in the new data that the model sees.
\subsection{Routing mask validation}
\begin{figure}
    \includegraphics[width=\linewidth]{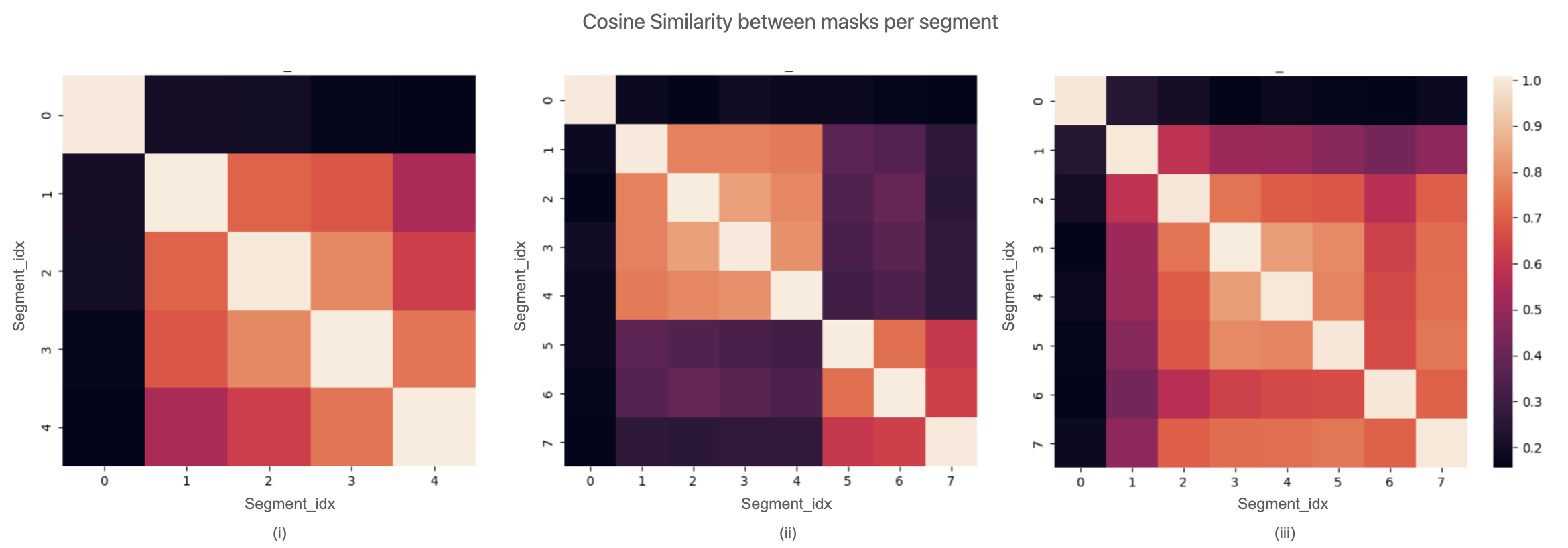}
    \caption{i: Starting from topic A and slowly transitioning to a slightly related topic B. ii: Starting from topic A and abruptly switching to unrelated topic B. iii: Starting from topic A, switching to unrelated topic B and coming back to topic A.}
    \label{fig:cosine_sim}
\end{figure}

We test the interpret-ability of the masks generated at the pre-training stage to validate that the expert usage pattern does not change very frequently but also change if the subject matter changes. To test this we compute the cosine similarity between the generated masks per segment for a couple of examples for 3 cases as shown in \cref{fig:cosine_sim}.

Through cosine similarities between the masks generated we can estimate the similarities in the experts used between each segment. Firstly, while processing inputs from a single topic it can be observed that the similarity between experts used per segment is very high. This can lead to high CHR. \cref{fig:cosine_sim}(i) also demonstrates that the similarity wears of as topic A transitions into topic B. Therefore, this model demonstrates adaptability to context change. Additionally, it can be also observed that when context switches drastically \cref{fig:cosine_sim}(ii) the model changes its experts quite drastically. Finally, \cref{fig:cosine_sim}(iii) demonstrates that similar experts are selected when conversation changes to an old topic after a short deviation to an unrelated topic.

\subsection{Fine-Grained Core-En Results}
\label{app:core-en}
Table~\ref{tab:core-en-breakdown} reports the task-level Core-En results for dense baselines and Flexible Stepped-MoE (Flexible S-MoE) with 12B total parameters at 1B and 3B active sizes. The Flexible S-MoE configurations use segment lengths $S=32$ and $S=16$, with each model trained and evaluated at its respective segment length. 

At $S=32$, Flexible S-MoE has lower Core-En-0s scores than the dense baselines, with particularly pronounced deficits on ARC-Challenge and ARC-Easy at the 1B active size. Reducing the segment length to $S=16$ improves the reported Core-En-0s score from 0.640 to 0.666 at 1B and from 0.701 to 0.722 at 3B, exceeding the corresponding dense scores of 0.662 and 0.704. The shorter-segment models also improve Core-En-1s at both active sizes. Improvements are not uniform across tasks: for example, WinoGrande remains below the dense baseline at both sizes.

A possible explanation is that approximately 70\% of the ARC prompts questions are very short. Under the pretraining evaluation protocol, experts are initially selected from the first-token representation; short prompts may therefore complete without a subsequent expert refresh informed by more context. The gains from shorter segments are consistent with this explanation.

\begin{table}[ht]
\centering
\caption{Fine-grained Core-En results. Flexible S-MoE has 12B total parameters. Each $S$ configuration is trained and evaluated at the indicated segment length. Composite scores are reproduced as reported.}
\label{tab:core-en-breakdown}
\begin{adjustbox}{max width=.88\textwidth}
\begin{tabular}{c|c|c|c|c|c|c}
\toprule
\multirow{2}{*} {\makecell{Task}}& \multicolumn{2}{c}{Dense} & \multicolumn{2}{c}{Flexible, $S=32$} & \multicolumn{2}{c}{Flexible, $S=16$} \\
\cmidrule(lr){2-3}\cmidrule(lr){4-5}\cmidrule(lr){6-7}
 & 1B & 3B & 1B & 3B & 1B & 3B \\
\midrule
ARC-Challenge & 0.4189 & 0.4838 & 0.3686 & 0.4625 & \textbf{0.4369} & \textbf{0.5128} \\
ARC-Easy & \textbf{0.7605} & 0.7997 & 0.6987 & 0.7912 & 0.7563 & \textbf{0.8207} \\
HellaSwag & 0.5043 & 0.5517 & 0.5129 & 0.5704 & \textbf{0.5202} & \textbf{0.5810} \\
WinoGrande & \textbf{0.6283} & \textbf{0.6803} & 0.6140 & 0.6559 & 0.6188 & 0.6725 \\
LAMBADA & 0.6144 & 0.6720 & 0.6346 & 0.6953 & \textbf{0.6427} & \textbf{0.7167} \\
SciQ & 0.9410 & 0.9550 & 0.9330 & 0.9580 & \textbf{0.9460} & \textbf{0.9610} \\
PiQA & \textbf{0.7639} & 0.7851 & 0.7149 & 0.7704 & 0.7454 & \textbf{0.7867} \\
\midrule
Core-En-0s & 0.662 & 0.704 & 0.640 & 0.701 & \textbf{0.667} & \textbf{0.722} \\
\midrule
TriviaQA & 0.2797 & 0.3789 & 0.2607 & 0.3725 & \textbf{0.3168} & \textbf{0.4042} \\
WebQS & 0.0999 & 0.1885 & 0.1240 & 0.1816 & \textbf{0.1299} & \textbf{0.1909} \\
\midrule
Core-En-1s & 0.190 & 0.284 & 0.192 & 0.277 & \textbf{0.223} & \textbf{0.298} \\
\bottomrule
\end{tabular}
\end{adjustbox}
\end{table}

\subsection{Compute-Normalized Training Comparison}
\label{app:compute-normalized}
To assess the effect of multi-budget training under a common approximate training-FLOP budget, we train a Flexible S-MoE model supporting two active sizes, 1B and 3B, on 1T unique tokens. Batch replication evaluates each sequence at both active sizes. We compare this shared checkpoint against individually trained Dense and Static S-MoE baselines, where Static S-MoE supports a single fixed active size. The 1B baselines are trained for two epochs over the 1T-token dataset, while the 3B baselines are trained on 667B tokens from the same dataset.

For normalization, we use the standard active-parameter approximation $C\approx 6NT$, where $N$ denotes active parameters and $T$ denotes processed training tokens. Processing 1T tokens at both 1B and 3B gives an approximate budget of $24\times10^{21}$ FLOPs for the flexible checkpoint. The two independently trained baseline models collectively use approximately the same budget: $12\times10^{21}$ FLOPs at each size. Thus, the comparison matches the combined approximate budget of the two-model baseline portfolio to that of the shared flexible model; each individual baseline receives approximately half of the flexible checkpoint's total budget. This approximation does not account for all routing, attention, communication, or hardware-utilization overheads.

Table~\ref{tab:compute-normalized} shows that Flexible S-MoE improves over Dense on MMLU-5s and GSM-8k at both active sizes, while scoring lower on Core-En-0s and Core-En-1s. Static S-MoE outperforms Flexible S-MoE on all four reported metrics at both sizes. The results illustrate a trade-off between supporting multiple active budgets in one checkpoint and optimizing a separate model for each budget.

\begin{table}[ht]
\centering
\caption{Compute-normalized results. The shared Flexible S-MoE checkpoint supports both active sizes. The combined approximate training-FLOP budget of the two baselines within each model family is matched to that of the flexible checkpoint.}
\label{tab:compute-normalized}
\begin{tabular}{c|c|c|c|c|c}
\toprule
Model & Active size & Core-En-0s & Core-En-1s & MMLU-5s & GSM-8k \\
\midrule
Dense & 1B & 0.664 & 0.192 & 0.476 & 0.179 \\
Dense & 3B & 0.692 & 0.278 & 0.570 & 0.341 \\
\midrule
Static S-MoE & 1B & 0.662 & 0.217 & 0.552 & 0.188 \\
Static S-MoE & 3B & 0.690 & 0.264 & 0.610 & 0.436 \\
\midrule
Flexible S-MoE & 1B & 0.650 & 0.191 & 0.498 & 0.181 \\
Flexible S-MoE & 3B & 0.686 & 0.262 & 0.602 & 0.425 \\
\bottomrule
\end{tabular}
\end{table}

The approximate FLOP normalization does not imply equal practical training cost. The reported training runs use 1,024 TPU cores for each Dense and Static S-MoE baseline and 2,048 TPU cores for the shared Flexible S-MoE model. Their respective reported costs are 30,720 and 29,123 TPU core-hours for Dense at 1B and 3B, 92,160 and 82,375 TPU core-hours for Static S-MoE, and 203,435 TPU core-hours for the flexible checkpoint. Although these measurements capture a practical training cost overhead beyond the active-parameter FLOP approximation the key benefit of using the Flexible S-MoE model lies in its ability to fit on limited storage while serving at multiple active sizes which the static S-MoE lacks.

\subsection{Cross-Segment-Length Evaluation}
\label{app:cross-segment}
We evaluate a single Flexible S-MoE checkpoint trained with segment length $S_{\mathrm{train}}=16$ using inference segment lengths $S_{\mathrm{eval}}\in\{16,32,64\}$. The model has 12B total parameters and operates at a 3B active size throughout. Unlike the comparison in Appendix~\ref{app:core-en}, these experiments change the inference segment length without retraining.

\begin{table}[htbp]
\centering
\caption{Cross-segment-length evaluation of a single Flexible S-MoE checkpoint with 3B active and 12B total parameters, trained at $S_{\mathrm{train}}=16$.}
\label{tab:cross-segment}
\begin{tabular}{rrrrr}
\toprule
$S_{\mathrm{eval}}$ & Core-En-0s & Core-En-1s & MMLU-5s & GSM-8k \\
\midrule
16 & 0.722 & 0.298 & 0.653 & 0.513 \\
32 & 0.706 & 0.284 & 0.654 & 0.508 \\
64 & 0.695 & 0.217 & 0.650 & 0.489 \\
\bottomrule
\end{tabular}
\end{table}

Increasing the inference segment length from 16 to 32 largely preserves performance. Longer segments reduce expert-refresh frequency, providing a potential mechanism for lowering I/O overhead at minimal cost to accuracy.


\end{document}